\documentclass[11pt]{article}
\usepackage[preprint]{acl}
\usepackage{times}
\usepackage{latexsym}
\usepackage[T1]{fontenc}
\usepackage[utf8]{inputenc}
\usepackage{microtype}
\usepackage{inconsolata}
\usepackage{graphicx}
\usepackage{amssymb}

\usepackage{makecell}
\usepackage{multirow}
\usepackage{pifont}
\usepackage[most,skins,theorems]{tcolorbox}
\usepackage{booktabs}
\usepackage{placeins}
\usepackage{tabularx}
\usepackage{array}
\usepackage[normalem]{ulem}
\usepackage{xcolor}
\usepackage{subcaption}

\usepackage{dblfloatfix}
\usepackage{listings}
\usepackage[most]{tcolorbox}

\tcbuselibrary{breakable,listings}
\newtcblisting{promptbox}[1]{
    enhanced,
    breakable,
    listing only,
    listing engine=listings,
    colback=gray!4,
    colframe=gray!45,
    boxrule=0.5pt,
    arc=1mm,
    left=1.5mm,
    right=1.5mm,
    top=1mm,
    bottom=1mm,
    title={#1},
    fonttitle=\bfseries,
    listing options={
        basicstyle=\ttfamily\footnotesize,
        breaklines=true,
        breakatwhitespace=false,
        columns=fullflexible,
        keepspaces=true,
        showstringspaces=false,
        upquote=true
    }
}

\newcommand{\skill}{SciDSK}

\title{Scientific Data Skills: Enabling Agent-Ready Scientific Data Services \\at Scale}

\author{
Xiaohan Huang\thanks{These authors contributed equally to this work.},
Qingqing Long\footnotemark[1],
Xiaolei Du,
Siyu Pu,
Jiawen Xu,
Haotian Chen,\\
\textbf{Chenyang Zhao,
Jinbiao Liu,
Xuezhi Wang,
Hengshu Zhu\thanks{Corresponding authors: Hengshu Zhu (\texttt{hszhu@cnic.cn}) and Yuanchun Zhou (\texttt{zyc@cnic.cn}).},
Yuanchun Zhou\footnotemark[2]} \\
Computer Network Information Center, Chinese Academy of Sciences \\
University of the Chinese Academy of Sciences\\
Beijing, China
}

\begin{document}
\maketitle

\begin{abstract}
Scientific data are increasingly used by AI agents, yet existing dataset representations provide limited support for reliable dataset discovery and interpretation, constraining their effective use in scientific workflows.
This limitation arises because agents must search across heterogeneous repositories and reconstruct dataset-specific semantics and operating procedures from documentation designed primarily for human use.
To address this limitation, we introduce the Scientific Data Skill (\skill{}), an agent-ready representation that packages dataset-specific knowledge and operational guidance as a reusable agent skill.
A \skill{} integrates dataset descriptions, scientific context, file organization, task-specific usage procedures, quality checks, and provenance information while retaining the underlying data in its original repository.
We define a structured \skill{} specification and develop a systematic construction pipeline that grounds each \skill{} in authoritative dataset records and associated supporting materials.
We further establish the Scientific Data Skill Bank, a unified platform that publishes \skill{} resources across six scientific disciplines and supports package access, persistent identification, and traceability to source datasets.
We evaluate \skill{} through a retrieval benchmark for dataset discovery and controlled cases for dataset interpretation.
On the query retrieval benchmark, Agent-SciDSK achieves 80.77\% Hit@1, exceeding Agent-Raw by 9.62 percentage points.
Across controlled interpretation cases, the \skill{} condition satisfies 23 of 24 assessment criteria, compared with 22 under the web-page condition.
These results indicate that \skill{} improves how agents locate and understand scientific datasets, providing a stronger foundation for actionable scientific data use.

\end{abstract}

\section{Introduction}

AI agents have gained increasing capabilities to plan tasks~\cite{hu2025agentgen,feng2026cocoa,luo2025large}, use external tools~\cite{doshi2026towards}, and execute multi-step workflows~\cite{zhang2025comprehend}, supported by recent advances in large language models (LLMs)~\cite{singh2025openai,xu2026deepseek, zhu2025can,yan2024inductive}.
Recent AI for Science (AI4S) advances have increasingly incorporated such agents into scientific reasoning, experimentation, and data analysis~\cite{xiang-etal-2026-llm,long2026survey, chen2025scirerankbench}.
In scientific research, such agents could accelerate data-intensive discovery~\cite{huang2026autonomous, hou2026bioflowbench,qin2025scihorizon,huang2026scihorizon}.
They can also assist researchers in identifying relevant datasets, understanding their contents, and incorporating them into computational analyses~\cite{viswanathan2023datafinder, long2026sciencedb, gao2025democratizing, hong-etal-2025-data}.
Realizing this potential requires scientific data representations tailored to agent-driven workflows.
Initiatives such as the FAIR principles have improved scientific data findability, accessibility, interoperability, and reusability~\cite{wilkinson2016fair, jacobsen2020fair}.
Yet scientific data resources remain distributed across heterogeneous repositories, and their accompanying documentation is organized primarily for human use~\cite{chapman2020dataset,batista2022machine,pushkarna2022data,song2020deep}.
Emerging AI-ready data approaches improve machine readability and programmatic access while supporting data-readiness assessment in machine-learning workflows~\cite{akhtar2024croissant,hiniduma2025data}.
However, these approaches provide limited support for the dataset-specific context and operational guidance required by AI agents.
Existing agent tools provide increasingly capable interfaces for searching repositories, calling APIs, and operating on files.
However, these tools do not generally provide the dataset-specific knowledge agents need to discover relevant datasets and interpret their files.
Agents must still reconstruct this knowledge from heterogeneous metadata records and human-oriented documentation.
This separation between executable operations and dataset-specific context creates two closely related challenges for scientific data use.

\begin{figure}[t]
\centering
\includegraphics[width=\linewidth]{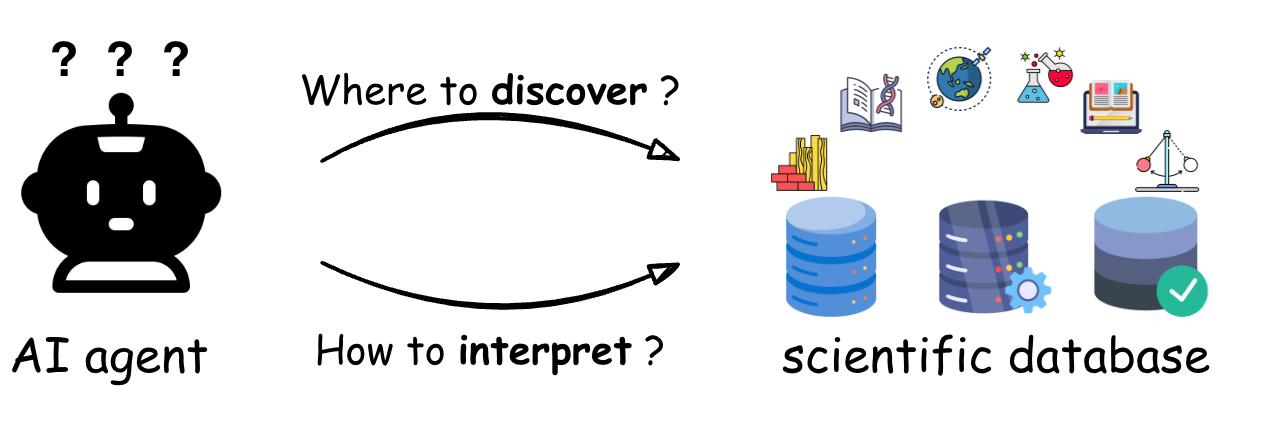}
\vspace{-7mm}
\caption{Key challenges in enabling AI agents to discover and interpret scientific data.}
\label{fig:intro_motivation}
\end{figure}

Scientific data therefore remain insufficiently agent-ready for reliable discovery and interpretation, as summarized in Figure~\ref{fig:intro_motivation}.
\textbf{First, agents must resolve scientific information needs into reliable dataset candidates across a fragmented data landscape.}
A scientific request rarely identifies which repository or data service is most likely to contain the needed dataset~\cite{chapman2020dataset}.
Agents must therefore route the request to candidate sources, reformulate it for repository-specific metadata and search interfaces, and compare results described with heterogeneous terminology.
Because these steps precede candidate evaluation, discovery becomes brittle when an agent lacks a unified, agent-readable search space.
This agent-side routing and retrieval burden impedes systematic discovery across sources~\cite{medina2021controlled,viswanathan2023datafinder}.

\textbf{Second, agents must infer dataset structure and file-level semantics from incomplete descriptions.}
Dataset-level metadata commonly summarizes a dataset's scientific scope, provenance, and general content~\cite{huang2025perceptual,hafner2025data,zhang2026cite}.
However, such metadata often omits file organization, the scientific roles of individual files, and the relationships among them~\cite{batista2022machine}.
File formats alone cannot resolve these semantics, because similar formats may encode different measurements or processing stages~\cite{walter2026fred}.
Consequently, agents may select inappropriate files, misinterpret their contents, or propagate unsupported assumptions into downstream analyses.
These two challenges directly constrain scientific data use.
An agent cannot reliably act on a dataset that it cannot locate or whose files and scientific semantics it does not understand.
Errors at either stage can propagate into data selection, preparation, and downstream analysis~\cite{chen2025scienceagentbench}.
Improving dataset discovery and interpretation is therefore a prerequisite for building actionable, closed-loop scientific data workflow capabilities~\cite{liu2024genotex}.
This motivates a central question: \textit{how can scientific datasets be represented so that agents can reliably discover and interpret them?}

Our key insight is that agent skills provide a modular mechanism for equipping AI agents with task-specific knowledge and operational guidance~\cite{xu2026agent}.
Through progressive disclosure, AI agents identify relevant skills from concise descriptions and load detailed instructions only when needed.
Building on this mechanism, we introduce the \textbf{Scientific Data Skill (\skill{})}, an agent-ready representation that organizes a scientific dataset's associated knowledge and usage procedures as a reusable agent skill.
A \skill{} organizes dataset descriptions, scientific context, file organization, operational guidance, usage constraints, and provenance within a unified skill package.
Once installed, a \skill{} makes its associated dataset discoverable to agents through the skill description, without requiring an initial query to the source repository.
After the \skill{} is selected, its detailed instructions explain the dataset's scientific scope, file roles, and structural relationships, thereby supporting file-level interpretation.
The instruction body also records documented usage procedures and quality checks that agents can consult when preparing the dataset for analysis.
By supporting reliable discovery and interpretation, \skill{} helps agents move from dataset identification toward more actionable, closed-loop scientific data workflows with available tools.

\begin{table*}[t]
\centering
\caption{Comparison of SciDSK with existing dataset- and agent-oriented concepts.}
\label{tab:concept_comparison}
\resizebox{0.9\linewidth}{!}{
\begin{tabular}{ccccc|c}
\toprule
 & Dataset Metadata & Dataset Card & Agent Skill & Tool/MCP & \textbf{SciDSK(Ours)} \\
\midrule
Dataset description
& \checkmark & \checkmark & $\times$ & $\times$ & \checkmark \\

Scientific context
& Partial & \checkmark & Partial & $\times$ & \checkmark \\

Task knowledge
& $\times$ & Partial & \checkmark & $\times$ & \checkmark \\

Operational guidance
& $\times$ & Partial & \checkmark & \checkmark & \checkmark \\

Agent discovery
& Partial & Partial & \checkmark & $\times$ & \checkmark \\

File-level interpretation
& Partial & Partial & Partial & $\times$ & \checkmark \\

\bottomrule
\end{tabular}
}
\end{table*}

We further establish the \textbf{Scientific Data Skill Bank} as an online platform for publishing and accessing a curated collection of \skill{} resources.
Resources in the current collection are manually selected and reviewed before publication against platform-defined criteria for source authenticity, representation fidelity, skill safety, and agent compatibility.
Each published \skill{} is assigned a CSTR~\cite{zhou2026construction} and records its source dataset's identifier and provenance metadata, allowing the two resources to be independently identified, cited, and traced.
The main contributions of this work are summarized as follows.

\noindent\textbf{(1)}
We introduce \skill{}, a reusable agent skill that makes dataset-specific knowledge and usage procedures agent-ready, together with a common specification and systematic construction pipeline.

\noindent\textbf{(2)}
We establish the Scientific Data Skill Bank, which publishes a curated collection of \skill{} resources linked to their source datasets and assigned individual CSTRs.

\noindent\textbf{(3)}
We conduct empirical evaluations of \skill{} across dataset discovery and interpretation.
The results indicate that \skill{}-based workflows can improve dataset discovery and support more precise interpretation.

\section{Related Work}

In this section, we review three lines of research related to Scientific Data Skills.
We first examine efforts to improve the AI readiness of scientific data through AI-readable representations, dataset documentation, and data-readiness frameworks.
We then discuss agent tools developed for scientific data retrieval, processing, and analysis.
Finally, we examine agent skills as reusable representations of knowledge and operational guidance, with particular attention to their limited support for individual scientific datasets.

\subsection{Scientific Data Representations Toward AI Readiness}
\sloppy
Scientific data representations have evolved from repository-oriented metadata toward richer descriptions~\cite{akhtar2024croissant,assante2016scientific}.
The FAIR principles, persistent identifiers, metadata standards, and FAIR Digital Objects improve dataset discovery, attribution, exchange, and reuse across data infrastructures~\cite{wilkinson2016fair,batista2022machine,de2020fair}.
Dataset documentation frameworks further describe collection processes, intended uses, limitations, and ethical considerations.
Formats such as RO-Crate~\cite{soiland2022packaging} and Croissant~\cite{akhtar2024croissant} provide structured representations of research objects, data resources, record structures, field semantics, provenance, and computational access.
Recent AI-readiness frameworks extend this scope to data quality, governance, sustainability, and task suitability~\cite{hiniduma2025data,clark2024ai,majithia2026actionable}.
Together, these efforts make scientific datasets more discoverable, interpretable, and accessible to computational systems.
However, existing representations primarily support data publication, assessment, exchange, or model-development pipelines~\cite{soiland2022packaging, akhtar2024croissant, brewer2026data}.
These descriptions are largely declarative, specifying what a dataset contains, how it was produced, and how its records can be accessed.
They do not generally organize the dataset-specific knowledge required for autonomous use, including when the dataset should be selected, how its scientific concepts correspond to particular files and fields, which preparation procedures should be applied, and how the resulting data should be validated.
Such knowledge often remains distributed across metadata records, documentation pages, loaders, and example workflows.
Consequently, a dataset may be FAIR, extensively documented, and programmatically accessible without being agent-ready for reliable, task-specific use by a general-purpose AI agent.
Bridging this gap requires an agent-ready representation that integrates dataset identity, scientific context, structural semantics, operational guidance, usage constraints, and validation procedures.

\subsection{Agent Tools for Scientific Data}
Agent tools extend AI agents with the ability to interact with external data services and computational environments~\cite{ding2025scitoolagent, wolflein-etal-2025-llm}.
Existing systems use such tools to retrieve records from scientific databases and repositories, process particular data formats, and invoke domain-specific analytical software~\cite{long2026sciencedb,gao2025democratizing,hong-etal-2025-data}.
These tools typically encapsulate bounded operations associated with a particular data source, format, or computational package~\cite{chaudhari2026modular, pham2026chemgraph}.
Their interfaces describe available functions, input parameters, and returned outputs, allowing agents to execute specialized procedures within scientific workflows~\cite{yuan-etal-2025-easytool, faghih2025tool}.
However, these operation-oriented interfaces do not generally organize knowledge around individual datasets.
Information about a dataset's scientific scope, file organization, file-level semantics, provenance, and usage constraints often remains distributed across metadata records and documentation.
Consequently, access to scientific tools does not ensure that an agent can identify a relevant dataset or reliably interpret its files.
Addressing this limitation requires a dataset-oriented representation that complements executable tools with dataset-specific knowledge and operational guidance.

\subsection{Agent Skills for Scientific Data}
Agent Skills provide a modular mechanism for extending AI agents with specialized knowledge, instructions, and reusable resources.
A skill typically exposes a concise description for discovery and provides detailed guidance that is loaded only when relevant.
This progressive disclosure allows agents to select appropriate capabilities without incorporating all instructions into their working context~\cite{anthropicclaudecodeskills,openaibuildskills}.
Existing skills support a wide range of tasks, including software development, document processing, data analysis, and domain-specific workflows~\cite{li2025skillflow}.
These examples demonstrate that procedural knowledge can be packaged independently of agent models and reused across tasks and runtime environments~\cite{li2026skillsbench,xu2026agent}.
However, existing agent skills are predominantly organized around tasks, tools, or general workflows rather than individual scientific datasets.
They do not typically define stable associations with dataset snapshots or systematically organize dataset-specific scientific context, file structure, provenance, and usage constraints.
Consequently, an agent may possess reusable data-analysis skills while still lacking the knowledge needed to identify and interpret a particular dataset.
Extending the Agent Skill paradigm to scientific data therefore requires a structured representation that connects each skill to a specific dataset and organizes the knowledge needed for reliable discovery and file-level interpretation.
Together, these capabilities provide the foundation for agents to apply dataset-specific operational guidance in downstream scientific workflows.

\section{Scientific Data Skill}

In this section, we present the conceptual foundation, representation specification, and construction pipeline of \skill{}.
We first explain how \skill{} extends the agent skill paradigm and relates to existing data representations and executable interfaces.
We then define its representation structure, dataset association and versioning scheme, and validation requirements.
Finally, we describe the pipeline for constructing \skill{} resources from scientific datasets and their supporting materials.

\subsection{Overview of Scientific Data Skills}
A \skill{} is an agent-ready representation of a scientific dataset that supports its discovery and file-level interpretation by AI agents.
It organizes dataset descriptions, file organization, operational guidance, and provenance information within a unified skill package.
By making a dataset easier to locate and understand, \skill{} provides a grounded basis for its downstream use in scientific workflows.
A \skill{} remains separate from its associated dataset and maintains an explicit link to the source dataset in its original repository.

\noindent \textbf{From Agent Skill to Scientific Data Skill.}
Agent skills package specialized knowledge, instructions, and resources as reusable capabilities for AI agents.
Following this paradigm, we extend the concept of skills from task-oriented agent capabilities to scientific data resources.
A \skill{} represents the dataset-specific knowledge and operational guidance required for agents to effectively interact with an associated dataset.
Rather than encapsulating the data itself, \skill{} makes the associated dataset discoverable and organizes the scientific and structural context needed for its interpretation.
This knowledge can subsequently guide agents as they prepare and use the dataset with available tools.

\noindent \textbf{\skill{} in Relation to Existing Concepts.}
Table~\ref{tab:concept_comparison} compares \skill{} with dataset metadata, dataset cards, agent skills, and tool interfaces.
Dataset metadata and dataset cards describe dataset characteristics, provenance, and intended uses.
Building on these representations, \skill{} organizes the scientific context and file-level information needed to support dataset discovery and interpretation.
Tools and MCP-based interfaces expose executable operations to AI agents.
The dataset-specific context and procedures provided by \skill{} help agents determine when and how to apply these operations.
In this way, \skill{} bridges descriptive data representations and executable interfaces for agent-driven use of scientific datasets.

\subsection{Scientific Data Skill Specification}
\begin{figure}[t]
\centering
\includegraphics[width=\linewidth]{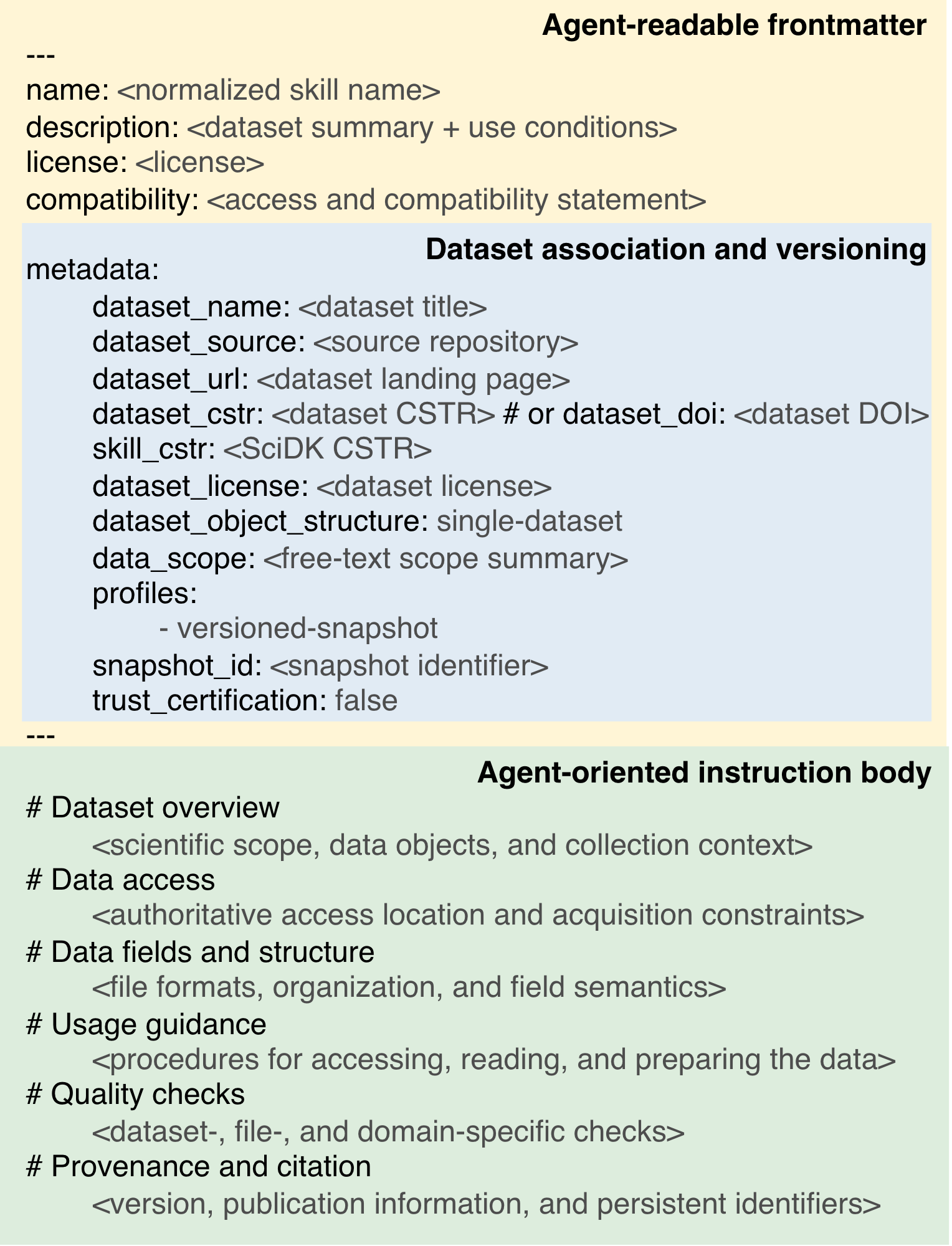}
\vspace{-5mm}
\caption{Schematic structure of a Scientific Data Skill.}
\label{fig:dataskill_schema}
\end{figure}

The \skill{} specification defines how the information needed for dataset discovery and interpretation is organized within a skill package.
Following existing agent skill conventions, each \skill{} uses a \texttt{SKILL.md} file as its core representation and may include additional resources when needed.
Within \texttt{SKILL.md}, YAML frontmatter exposes agent-readable descriptors for discovering the associated dataset and recording its identity and provenance.
The Markdown body provides the scientific and structural context needed to interpret the dataset.
It also records operational guidance that can support subsequent data preparation and use.
Figure~\ref{fig:dataskill_schema} illustrates this two-part structure.

\noindent\textbf{Agent-readable frontmatter.}
The YAML frontmatter contains agent-readable descriptors for dataset discovery, association, versioning, and provenance tracking.
Its top-level fields follow existing agent skill conventions and include \texttt{name}, \texttt{description}, \texttt{license}, and \texttt{compatibility}.
The nested \texttt{metadata} block adds dataset-specific descriptors, including dataset identity, source and access location, persistent identifiers, license, data scope, object structure, and snapshot information.
The \texttt{description} field summarizes the associated dataset and indicates when the \skill{} should be selected, thereby serving as its primary routing signal.
Together, these descriptors allow agents to assess the relevance of the associated dataset and identify its source and represented version.

\noindent\textbf{Agent-ready instruction body.}
The Markdown body is loaded after a \skill{} has been selected and provides the information needed to work with the associated dataset.
It comprises six components: dataset overview, data access, data fields and structure, usage guidance, quality checks, and provenance and citation.
The dataset overview, data fields and structure components describe the scientific context, file organization, data formats, and field semantics needed for interpretation.
The data access, usage guidance, and quality checks components specify documented procedures for obtaining, reading, preparing, and checking the data.
The provenance and citation component records persistent identifiers, version information, and publication details for traceability.
Together, these components support file-level interpretation and provide grounded guidance for subsequent data preparation with available tools.

\noindent\textbf{Dataset association and versioning.}
Each \skill{} represents a single scientific dataset, while the underlying data remain in their original repository.
The association is recorded through the dataset source, landing-page URL, and persistent identifier.
Separate identifier fields distinguish the dataset from its corresponding \skill{} and make their relationship traceable.
The \texttt{versioned-snapshot} profile indicates that the \skill{} describes a specific dataset snapshot identified by \texttt{snapshot\_id}.
Detailed version and publication information in the instruction body further identifies the dataset state to which the instructions apply.
Together, these records maintain an explicit relationship between the \skill{} and the represented dataset snapshot.

\noindent\textbf{Validation requirements.}
Before publication, each \skill{} undergoes checks for structural conformance, consistency with its source materials, and package integrity.
Structural checks cover the YAML frontmatter, required instruction components, and package organization.
Source-consistency checks compare dataset identifiers, access information, licensing information, file structure, and version records with the associated dataset materials.
Package-integrity checks confirm that the distributed \skill{} can be parsed and installed according to the skill conventions.
These checks assess the conformance and traceability of the \skill{} representation.

\subsection{Skill Construction Pipeline}
\begin{figure}[t]
\centering
\includegraphics[width=\linewidth]{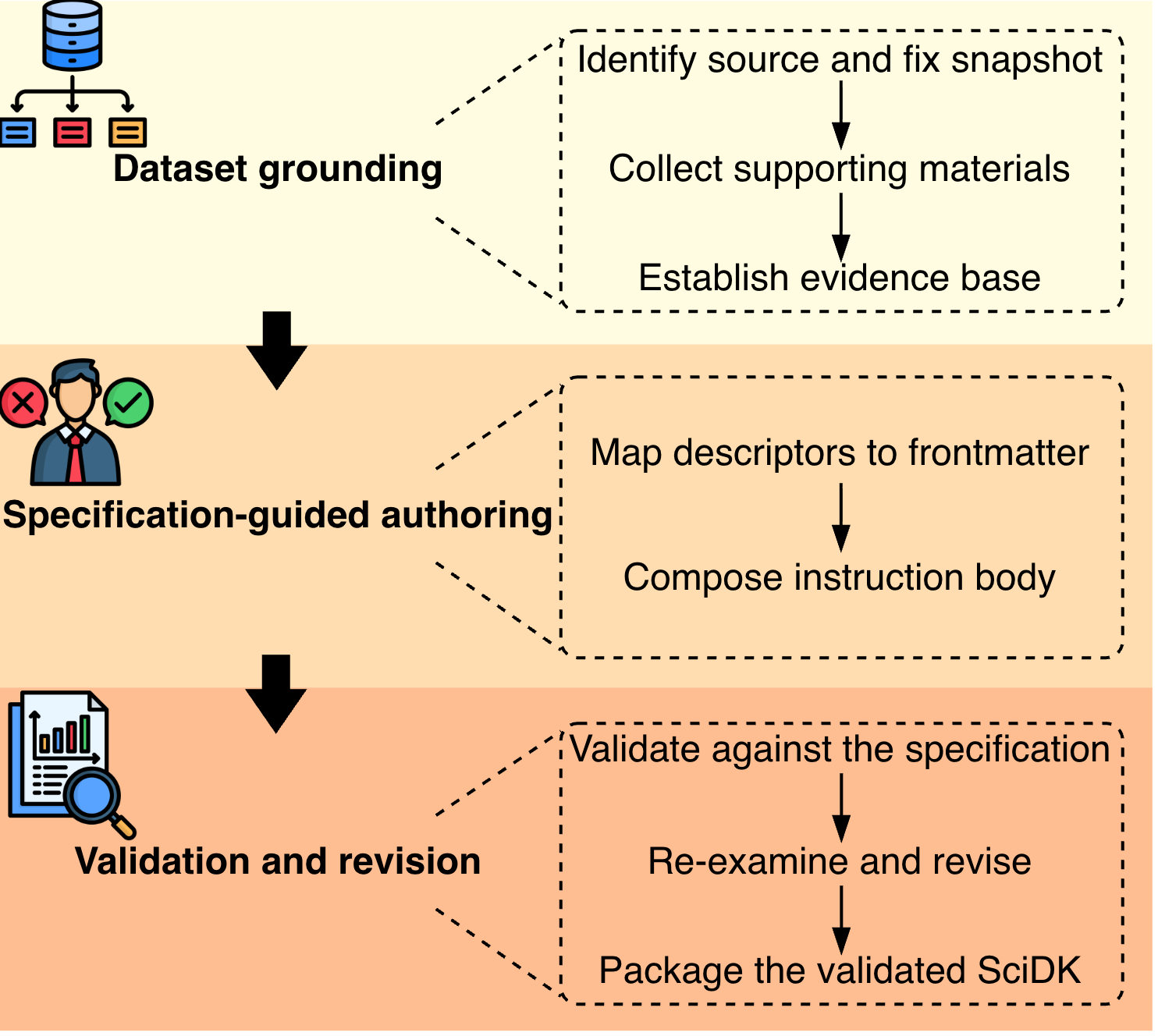}
\vspace{-6mm}
\caption{Construction pipeline of a Scientific Data Skill.}
\label{fig:dataskill_pipeline}
\end{figure}

The \skill{} construction pipeline provides a structured process for representing a scientific dataset, as shown in Figure~\ref{fig:dataskill_pipeline}.
Starting from a scientific dataset and its supporting materials, the pipeline establishes a traceable evidence base, organizes the information needed for dataset discovery and interpretation, incorporates documented guidance for downstream use, and checks the resulting representation through iterative revision.
The pipeline comprises three stages: dataset grounding, specification-guided authoring, and validation and revision.

\begin{figure*}[htbp]
    \centering
    \begin{subfigure}[t]{0.32\textwidth}
        \centering
        \includegraphics[width=\linewidth]{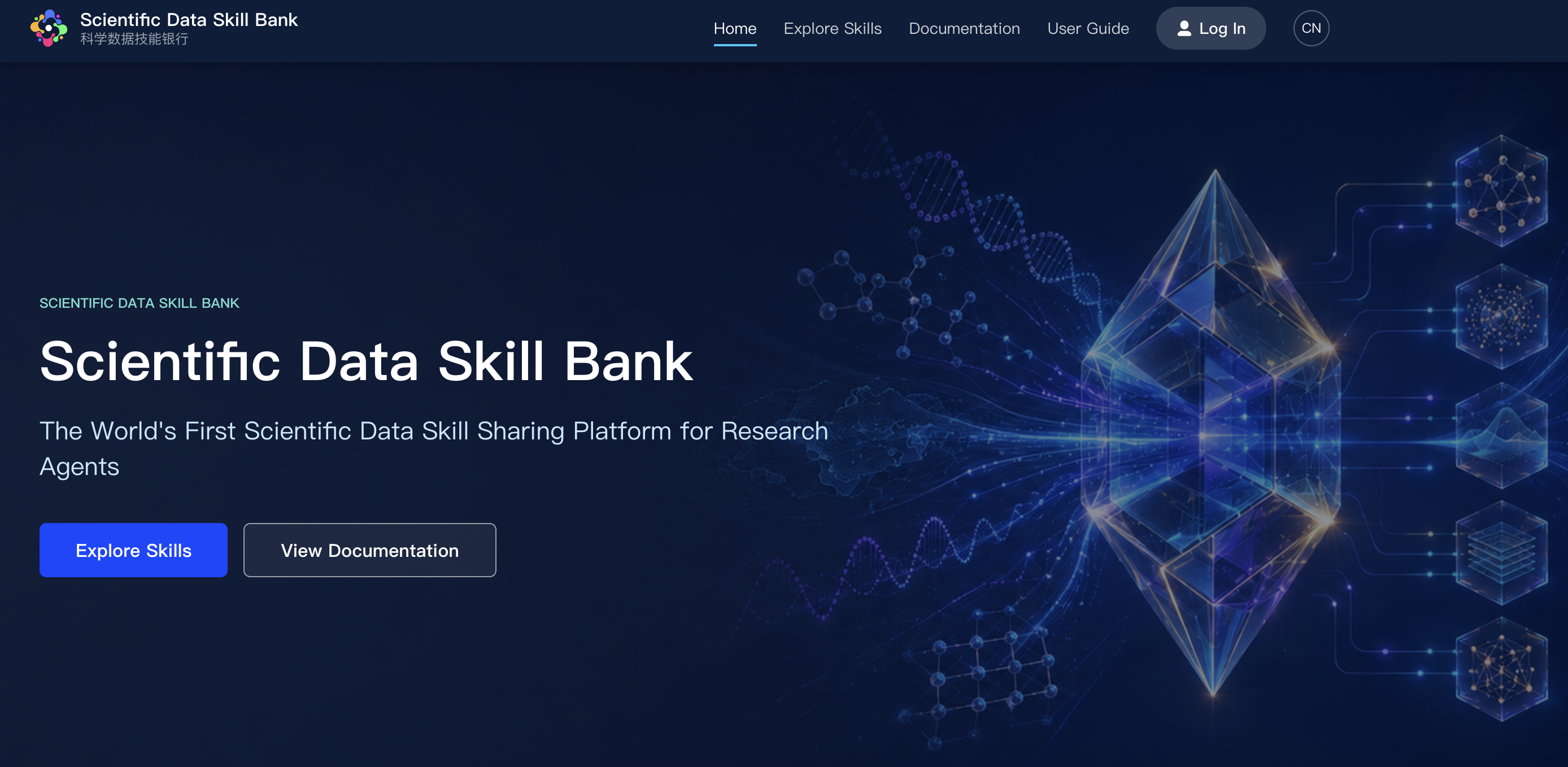}
        \caption{Platform Homepage.}
        \label{fig:platform_homepage}
    \end{subfigure}
    \hfill
    \begin{subfigure}[t]{0.32\textwidth}
        \centering
        \includegraphics[width=\linewidth]{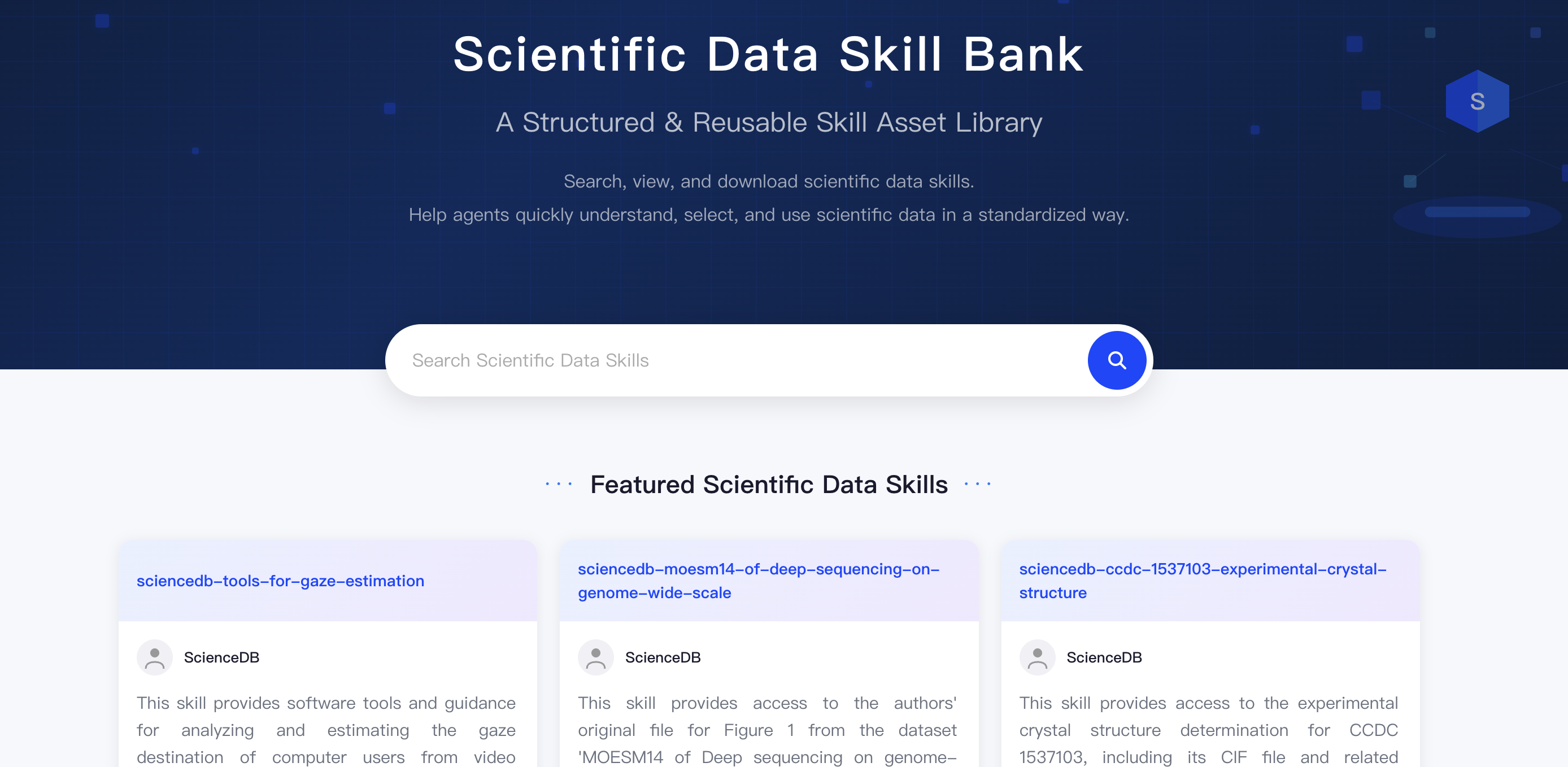}
        \caption{Skill Discovery Page.}
        \label{fig:skill_find}
    \end{subfigure}
    \hfill
    \begin{subfigure}[t]{0.32\textwidth}
        \centering
        \includegraphics[width=\linewidth]{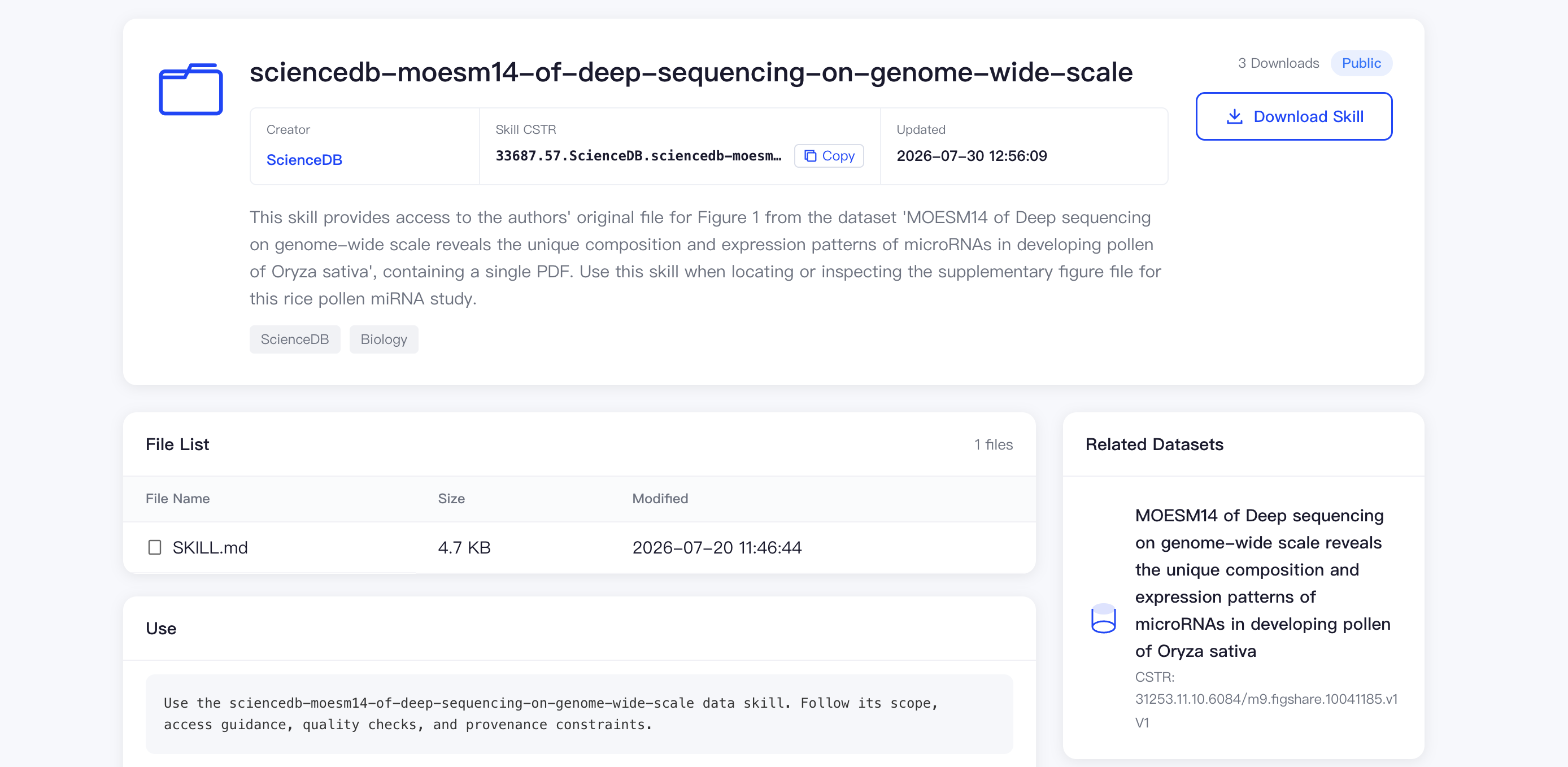}
        \caption{Data Skill Detail Information.}
        \label{fig:skill_detail}
    \end{subfigure}
    \vspace{-2mm}
    \caption{The online Scientific Data Skill Bank, which can be visited at  \url{https://scidsk.cn/}.}
    \label{fig:platform_overview}
\end{figure*}

\noindent\textbf{Dataset grounding.}
Dataset grounding defines the dataset and establishes the evidence base from which a \skill{} is constructed.
The process first identifies the authoritative dataset source and the specific dataset snapshot to be represented.
Available supporting materials are then collected, including dataset metadata, landing pages, documentation, file inventories, data dictionaries, associated publications, access conditions, licenses, and version records.
Information from these materials is normalized while its source attribution is preserved.
Missing or inconsistent information is recorded for subsequent review.
The resulting evidence base constrains the dataset-specific facts and usage knowledge that can be included in the \skill{}.

\noindent\textbf{Specification-guided authoring.}
Specification-guided authoring organizes the grounded evidence according to the \skill{} specification.
Dataset identity, access, licensing, versioning, and provenance information are mapped to the agent-readable frontmatter.
Scientific scope, data organization, field semantics, access procedures, usage guidance, and quality checks are organized within the corresponding components of the agent-ready instruction body.
Operational guidance is derived from the documented characteristics and constraints of the dataset.
Unsupported information is omitted, while required information that cannot be established from the available evidence is marked as unavailable.

\noindent\textbf{Validation and revision.}
The authored \skill{} is checked against the structural and content requirements of the \skill{} specification.
These checks cover the conformance of \texttt{SKILL.md} and its optional resources, their consistency with the grounded evidence, and the internal consistency between the frontmatter and instruction body.
When missing, contradictory, or unsupported information is identified, the source materials are re-examined and the authored content is revised accordingly.
The review and revision cycle continues until the identified issues have been addressed and the specification requirements are satisfied.
The finalized \texttt{SKILL.md} and any necessary supplementary resources are then organized into a \skill{} package associated with the corresponding dataset snapshot.
Package integrity is checked before publication.

\section{Scientific Data Skill Bank}
In this section, we present the Scientific Data Skill Bank, an online platform for publishing, discovering, and accessing a curated collection of \skill{} resources.
We first describe the platform interfaces and its initial cross-disciplinary resource collection.
We then outline the platform-defined review process applied before resource publication.
Finally, we explain how the platform supports resource discovery, package access, and traceability through persistent identifiers and explicit dataset associations.

\subsection{Platform Overview}
The Scientific Data Skill Bank\footnote{\url{https://scidsk.cn}} is an online platform for publishing, discovering, and accessing a curated collection of \skill{} resources.
The collection spans six disciplines: physics, chemistry, earth sciences, biology, materials science, and computer science and technology.
As shown in Figure~\ref{fig:platform_overview}, the platform provides three main interfaces for exploring the collection.
The homepage introduces the platform and provides an entry point to its published resources (Figure~\ref{fig:platform_homepage}).
The resource discovery interface supports keyword search and browsing by discipline and presents featured \skill{} resources (Figure~\ref{fig:skill_find}).
The detail page presents the content of an individual \skill{}, its association with the source dataset, its usage guidance, and a downloadable skill package (Figure~\ref{fig:skill_detail}).
Together, these interfaces allow users to browse, inspect, and download published \skill{} resources.

\subsection{Pre-publication Resource Review}
Before publication, each \skill{} is reviewed against its source materials and the proposed specification.
The platform defines four review dimensions: source authenticity, representation fidelity, skill safety, and agent compatibility.
Under source authenticity, the dataset identifier, source repository, license, and version information are compared with the records provided by the identified data source.
Representation fidelity is assessed by comparing the dataset description, file organization, access instructions, usage guidance, and provenance information with the available dataset records and supporting materials.
Skill safety review examines the package structure and included resources for evident risks, such as instructions that request unintended operations.
Agent compatibility review is limited to conformance with the expected agent skill structure and the parseability of the frontmatter and instruction body.
Only resources that meet the platform-defined publication criteria are included in the Scientific Data Skill Bank.
This review supports an internal publication decision.

\subsection{Resource Discovery, Access, and Traceability}

The platform supports resource discovery through keyword search, browsing by discipline, and featured entries.
Search and browsing results lead to detail pages where users can inspect the scientific scope, source dataset association, and usage information of individual \skill{} resources.
These pages allow users to assess resource relevance before accessing the corresponding package.

Each published \skill{} can be downloaded as a compressed skill package.
The underlying dataset is not distributed through the platform and remains accessible from its original repository.
Its access location and relevant instructions are recorded in the corresponding \skill{}.

Each published \skill{} is assigned an independent CSTR~\cite{zhou2026construction}.
Its frontmatter also records the CSTR of the associated dataset, or its DOI when a CSTR is unavailable, together with the snapshot identifier, source repository, and landing page.
These identifiers and source records connect the published \skill{} to the specific dataset snapshot described by its instructions.
The \skill{} and its associated dataset can therefore be independently identified and cited through their respective persistent identifiers.

\section{Evaluation Benchmark Construction}

In this section, we describe the construction of the tasks used to evaluate \skill{} in dataset discovery and interpretation.
We first present the retrieval benchmark for dataset discovery, including target dataset selection, query construction, and candidate corpus organization.
We then introduce the controlled cases for dataset interpretation and define their task requirements.

\subsection{Dataset Discovery Benchmark}
\label{sec:dataset_discovery}
\begin{figure}[h]
\centering
\includegraphics[width=\linewidth]{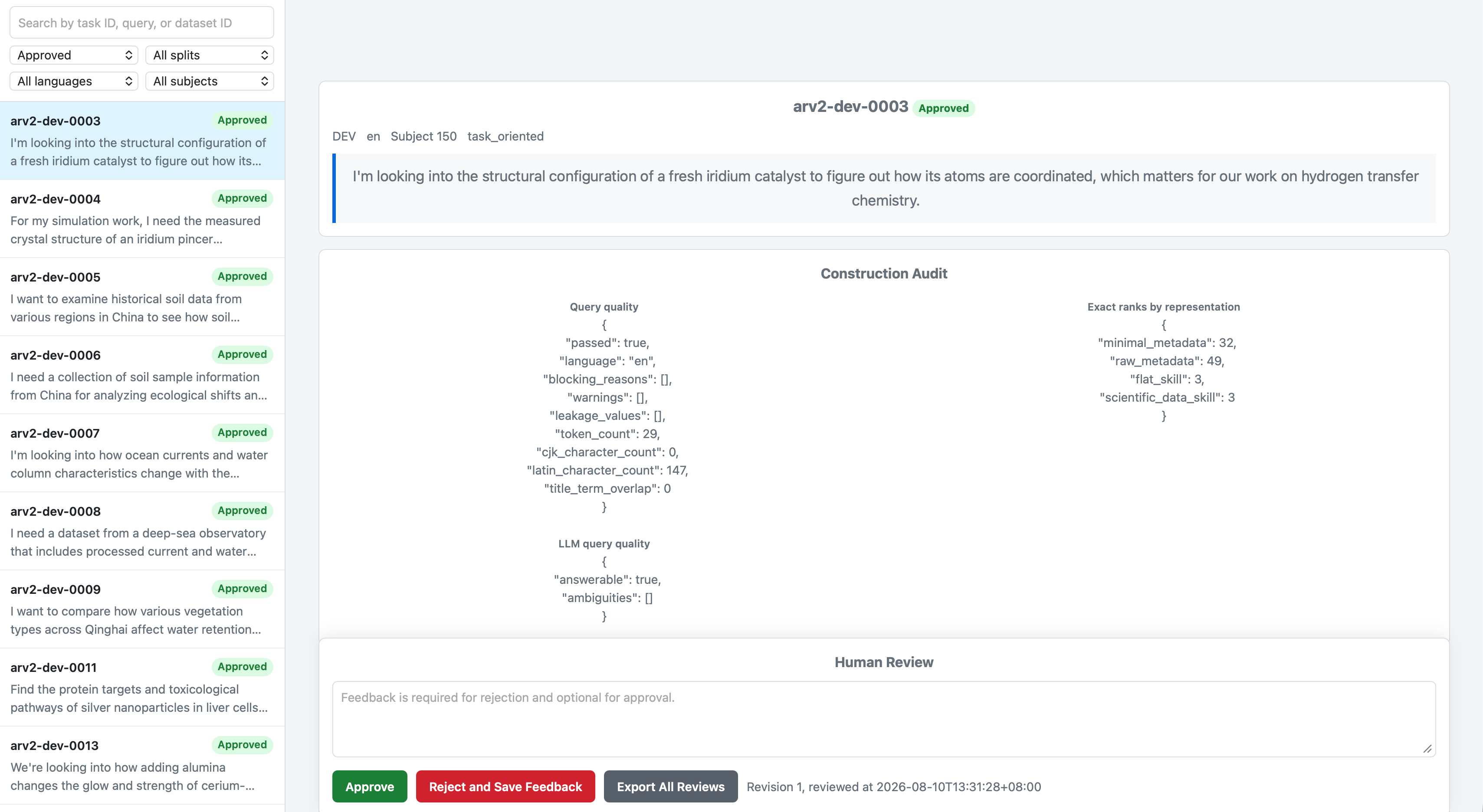}
\caption{Human annotation interface for reviewing discovery benchmark queries.}
\label{fig:discovery_annotation_platform}
\vspace{-3mm}
\end{figure}
We construct a retrieval benchmark using 72 datasets selected from six scientific disciplines.
For each target dataset, we construct two queries expressing research needs at different levels of specificity while excluding explicit identifying information.
This process yields 24 development queries and 120 initial test queries.
We further develop a human annotation platform to review all queries for naturalness, factual consistency, and target relevance.
Figure~\ref{fig:discovery_annotation_platform} shows the annotation interface.
After manual review, we exclude 16 ambiguous test queries, resulting in a final test set of 104 queries with one verified relevant dataset each.
The retrieval corpus covers the same six disciplines and is approximately four times the size of the public \skill{} collection.
For each candidate dataset, the corpus contains both a conventional dataset record and its corresponding \skill{} representation.
This one-to-one correspondence ensures that the compared methods retrieve and rank the same underlying dataset identities.

\subsection{Dataset Interpretation Cases}

We construct cases to assess whether \skill{} provides more precise and actionable support for dataset interpretation and select four representative cases for evaluation.
The selected cases cover CT image sequences, GIS rasters and sidecar files, image-based scientific tables, and cross-file associations between event labels and textual content.
Each case pairs a target dataset with a realistic request requiring the agent to explain the data content, identify file roles and organization, specify prerequisite checks, and distinguish evidence-supported information from the provided materials.
Dataset discovery and downstream analysis are excluded to isolate dataset interpretation.
Each case includes six atomic assessment criteria derived from the frozen source record and file tree.

\section{Experiment}

In this section, we evaluate \skill{} in dataset discovery and interpretation.
We first describe the experimental environment and agent configuration.
We then introduce the compared retrieval methods and evaluation metrics for dataset discovery.
Finally, we present the evidence conditions and evaluation procedure for dataset interpretation.

\begin{table}[t]
    \centering
    \caption{Overall discovery performance on the test sets. All values are percentages, and the best result is shown in bold.}
    \vspace{-2mm}
    \label{tab:discovery_overall}
    \resizebox{\linewidth}{!}{
    \begin{tabular}{ccccc}
        \toprule
        \textbf{Method} & \textbf{Hit@1} & \textbf{Recall@5} & \textbf{MRR} & \textbf{nDCG@5} \\
        \midrule
        BM25-Raw          & 47.12 & 69.23 & 57.59 & 59.12 \\
        Agent-Raw        & 71.15 & 90.38 & 79.04 & 81.90 \\
        Agent-SciDSK-Text  & 70.19 & 90.38 & 79.01 & 81.92 \\
        \midrule
        \textbf{Agent-SciDSK}       & \textbf{80.77} & \textbf{94.23} & \textbf{86.41} & \textbf{88.40} \\
        \bottomrule
    \end{tabular}
    }
    
\end{table}

\subsection{Experimental Setup}
\paragraph{Basic Settings}
All agent-based evaluations were conducted using a real agent environment as the harness, with Qwen3.6-Plus~\cite{yang2025qwen3} as the underlying language model.
The model's thinking mode was disabled, automatic tool selection was enabled, and the temperature was set to \(0\).
All other inference parameters remained at their default values.
Each model request had a timeout of 120 seconds, with up to three retries upon failure.
We used the July 2026 snapshot of the Scientific Data Skill Bank.
To maintain a controlled action space, we implemented task-specific tools based on the Model Context Protocol and exposed them through a strict allowlist.
The native Shell, general file-system, browser, and Web-search capabilities of the agent harness were disabled.
Session state was not retained across experimental runs.

\begin{table}[t]
\centering
\caption{Results on the four dataset interpretation cases.}
\label{tab:interpret_overall}
\resizebox{\linewidth}{!}{
\begin{tabular}{lcc}
\toprule
\textbf{Evidence condition} &
\textbf{Coverage} (\%) &
\textbf{Satisfied criteria} \\
\midrule
Web Page & 91.67 & 22/24 \\
\midrule
\textbf{Agent-SciDSK} & \textbf{95.83} & \textbf{23/24} \\
\bottomrule
\end{tabular}
}
\end{table}
\begin{table*}[!t]
\centering
\caption{Comparison on the CT skull reconstruction case.}
\vspace{-2mm}
\label{tab:interpret_ct_case}
\small
\renewcommand{\arraystretch}{1.35}
\setlength{\tabcolsep}{5pt}

\begingroup
\renewcommand{\tabularxcolumn}[1]{m{#1}}

\begin{tabularx}{\textwidth}{
    >{\raggedright\arraybackslash}X
    >{\raggedright\arraybackslash}X
    >{\raggedright\arraybackslash}X
    >{\raggedright\arraybackslash}X
    >{\centering\arraybackslash}m{1.3cm}
}
\toprule
\textbf{Evidence condition} &
\textbf{TIFF sequence interpretation} &
\textbf{Reported file-count handling} &
\textbf{Pre-use checks} &
\textbf{Coverage} \\
\hline

Dataset information page &
Incorrectly described the visible sequence as ``200+ slices'' &
Noted that the visible file tree may be incomplete &
Provided general checks for the image sequence and parameters &
5/6 \\
\hline

\textbf{Scientific Data Skill} &
\textbf{Correctly identified 196 consecutive slices} &
\textbf{Distinguished the reported total from the visible portion while preserving uncertainty} &
\textbf{Specified continuity, non-empty-file, readability, parameter-file, and directory-completeness checks} &
\textbf{6/6} \\
\bottomrule
\end{tabularx}
\endgroup
\end{table*}
\begin{table*}[!t]
\centering
\caption{Comparison on the Weibo rumor-event mapping case.}
\vspace{-2mm}
\label{tab:interpret_weibo_case}
\small
\renewcommand{\arraystretch}{1.35}
\setlength{\tabcolsep}{5pt}

\begingroup
\renewcommand{\tabularxcolumn}[1]{m{#1}}
\begin{tabularx}{\textwidth}{
    >{\raggedright\arraybackslash}X
    >{\raggedright\arraybackslash}X
    >{\raggedright\arraybackslash}X
    >{\raggedright\arraybackslash}X
    >{\centering\arraybackslash}m{1.3cm}
}

\toprule
\textbf{Evidence condition} &
\textbf{File organization} &
\textbf{Cross-file relationship} &
\textbf{Validation checks} &
\textbf{Coverage} \\
\hline

Dataset information page &
Identified \texttt{events.txt} as the label file and \texttt{posts.zip} as containing event-organized JSON content &
Linked event identifiers to event-named JSON files but left corpus-wide completeness unresolved &
Suggested general coverage and schema checks without explicit count and label-domain validation &
5/6 \\
\hline

\textbf{Scientific Data Skill} &
\textbf{Identified 4,664 labeled events and clearly distinguished the roles of the two files} &
\textbf{Specified a one-to-one mapping between event records and event-named JSON files} &
\textbf{Required count validation, binary-label checks, archive extraction, orphan detection, and sampled content verification} &
\textbf{6/6} \\
\bottomrule
\end{tabularx}
\endgroup
\end{table*}

\paragraph{Dataset Discovery Evaluation Protocols}
We compare four methods spanning lexical retrieval, agent-based document retrieval, and end-to-end \skill{} discovery.
\textbf{BM25-Raw} indexes conventional dataset records containing metadata and file-tree information.
\textbf{Agent-Raw} uses the same conventional dataset records but allows the agent to iteratively search, inspect candidates, and produce a final ranking.
\textbf{Agent-SciDSK-Text} follows the same agent-based retrieval protocol while treating complete \texttt{SKILL.md} documents as ordinary searchable text.
\textbf{Agent-SciDSK} first selects the two most relevant disciplines and then searches the registered \skill{} resources within the selected disciplines.
The search index uses the conventional metadata representation associated with each registered \skill{}.
All three agent-based methods use the same search budget and final-ranking protocol.
Each method produces a top-five ranking for every test query.
The agent-based methods are allowed at most four searches and ten candidate inspections per query and must construct their final rankings from previously retrieved candidates.
Both BM25 methods use \(k_1=1.5\) and \(b=0.75\), with ties resolved by dataset identifier.
We report Hit@1, Recall@5, mean reciprocal rank (MRR), and normalized discounted cumulative gain at rank five (nDCG@5).
The metric definitions are provided in Appendix~\ref{app:discovery_metrics}.

\paragraph{Dataset Interpretation Evaluation Protocols}
We compare three evidence conditions under the same agent framework.
\textbf{Web-Page} accesses a frozen snapshot of the corresponding ScienceDB landing page.
\textbf{Agent-SciDSK} accesses the complete \texttt{SKILL.md} through the native Agent Skill mechanism.
All conditions use the same model, request, prompt template, and runtime constraints, with external information sources disabled.
Each evidence condition is evaluated against 24 atomic criteria, comprising six criteria for each of the four cases.
Predefined term-matching rules determine whether each criterion is satisfied.
A single evaluator, blinded to the evidence condition, reviews every response for factual accuracy, unsupported inference, uncertainty handling, and potential matching errors.
The deterministic scores are retained for quantitative comparison, while discrepancies identified during review are reported separately.
We report the number of satisfied criteria out of 24, together with protocol compliance, tool calls, token usage, and runtime.
The complete assessment criteria are provided in Appendix~\ref{app:interpret_criteria}.

\label{sec:discovery_results}

\subsection{Dataset Discovery}

In this experiment, we answer the question: \textit{Does the end-to-end \skill{} workflow improve scientific dataset discovery over retrieval based on conventional records or static \skill{} documents?}
As shown in Table~\ref{tab:discovery_overall}, Agent-SciDSK achieves the best performance across all reported metrics.
The comparison between BM25-Raw and Agent-Raw isolates the effect of agent-based retrieval over the same conventional dataset records.
The substantial improvement of Agent-Raw demonstrates the value of iterative query formulation and candidate inspection over direct lexical retrieval.
The comparison between Agent-Raw and Agent-SciDSK-Text then changes the document representation while retaining the same agent-based retrieval protocol.
Their nearly identical performance indicates that treating complete \skill{} documents as ordinary searchable text provides little additional benefit over conventional records.
The comparison between Agent-SciDSK-Text and Agent-SciDSK examines the effect of using \skill{} as registered and routable agent skills.
Agent-SciDSK consistently improves all four retrieval metrics.
This result attributes the main performance gain to the end-to-end use of \skill{} within the agent workflow.

\subsection{Dataset Interpretation}

In this experiment, we answer the question: \textit{Does \skill{} provide more precise and actionable support for dataset interpretation?}
As shown in Table~\ref{tab:interpret_overall}, Agent-SciDSK achieves higher overall coverage, satisfying 23 of the 24 assessment criteria compared with 22 under the ScienceDB page condition.
The CT reconstruction case in Table~\ref{tab:interpret_ct_case} demonstrates the difference in quantitative interpretation and pre-use guidance.
The ScienceDB page condition incorrectly describes the visible TIFF sequence as containing more than 200 slices.
Agent-SciDSK correctly identifies 196 consecutive slices and distinguishes them from the reported dataset-level total of 1,576 files.
It also specifies checks for sequence continuity, file readability, parameter files, and directory completeness.
The Weibo event-mapping case in Table~\ref{tab:interpret_weibo_case} demonstrates the difference in cross-file interpretation.
Both conditions identify the roles of \texttt{events.txt} and \texttt{posts.zip}, but Agent-SciDSK more explicitly describes the correspondence between event records and event-named JSON files.
It further specifies checks for record and JSON-file counts, binary labels, archive extraction, orphaned identifiers, and sampled content consistency.
The ScienceDB page condition provides only general consistency checks without fully specifying these corpus-wide validations.

\section{Conclusion}

We presented \skill{}, an agent-ready representation that organizes dataset-specific knowledge and operational guidance for scientific dataset discovery and file-level interpretation.
By helping agents locate relevant datasets and understand their scientific and structural context, \skill{} provides a foundation for more actionable, closed-loop scientific data workflows.
We also developed its representation specification, construction pipeline, and publication platform.
The retrieval experiment shows that the end-to-end \skill{} workflow improves dataset discovery, while the controlled cases provide preliminary evidence of more precise file-level interpretation.

In the future, we will investigate version-aware maintenance of \skill{} resources as their source datasets evolve.This work will examine how changes in data files, metadata, documentation, and access conditions can trigger the construction and review of successor \skill{} versions. Each version should remain associated with its corresponding dataset snapshot, while links between successive versions preserve provenance and support reproducibility. We will also evaluate and adapt the \skill{} specification and construction pipeline for other heterogeneous data sources, with particular attention to differences in metadata schemas, persistent identifiers, documentation practices, and access mechanisms.

\section*{Acknowledgements}
We appreciate the contributions of the following individuals for their support of platform development: Chengzan Li, Jia Liu, Zeyu Zhang, Jidong Li, and Shu Wang.

\bibliography{cite}

\newpage
\appendix

\section{Metrics}
\label{app:discovery_metrics}
Let \(N\) denote the number of queries and \(r_i\) the rank of the target dataset for query \(i\).
We set \(r_i=\infty\) if the target is absent from the returned list.

\noindent\textbf{Hit@1.}
Hit@1 measures the proportion of queries for which the target dataset is ranked first:
\[
\mathrm{Hit@1}
=
\frac{1}{N}
\sum_{i=1}^{N}
\mathbb{I}(r_i=1).
\]

\noindent\textbf{Recall@5.}
Recall@5 measures whether the target dataset appears among the top five results:
\[
\mathrm{Recall@5}
=
\frac{1}{N}
\sum_{i=1}^{N}
\mathbb{I}(r_i\leq 5).
\]

\noindent\textbf{Mean Reciprocal Rank.}
MRR rewards methods that rank the target dataset earlier.
Ranks beyond the submitted top-five list receive a score of zero:
\[
\mathrm{MRR}
=
\frac{1}{N}
\sum_{i=1}^{N}
\frac{\mathbb{I}(r_i\leq 5)}{r_i}.
\]

\noindent\textbf{nDCG@5.}
With one relevant dataset per query, nDCG@5 is defined as:
\[
\mathrm{nDCG@5}
=
\frac{1}{N}
\sum_{i=1}^{N}
\frac{\mathbb{I}(r_i\leq 5)}
{\log_2(r_i+1)}.
\]
It measures ranking quality with a logarithmic discount for lower positions.

\noindent\textbf{Agent diagnostics.}
Search coverage is the proportion of queries for which the target dataset appears in the results returned by the search tools.
Discipline-routing Recall@2 is the proportion for which the target dataset's discipline is included among the two selected disciplines.
Submission rate is the proportion of queries that produce a nonempty ranked list.
Protocol violation rate is the proportion of queries containing an invalid dataset identifier or another violation of the output protocol.
Tool calls, token usage, and runtime are reported as per-query averages.

\section{Interpretation Assessment Criteria}
\label{app:interpret_criteria}

Each interpretation case is evaluated against six case-specific atomic criteria.
The criteria represent factual claims or usage guidance that should be present in a complete interpretation.
Predefined phrase-matching rules produce preliminary criterion-level scores, and rubric coverage is calculated as the proportion of matched criteria.
A single evaluator subsequently reviews every response without access to its evidence-condition label.
The review assesses factual accuracy, unsupported inference, uncertainty handling, and potential errors in the deterministic matching results.
The reported quantitative coverage retains the deterministic scores, while discrepancies identified through blinded review are reported separately in the results.

\paragraph{CT skull reconstruction.}
The assessment examined whether the response:
(i) identified the dataset as skull micro-CT data;
(ii) identified the TIFF stack and \texttt{params.ini} as relevant inputs;
(iii) recognized that the visible sequence from \texttt{0000} to \texttt{0195} contains 196 slices;
(iv) distinguished the reported 1,576 files from the visible TIFF sequence;
(v) recommended checks for sequence continuity, readability, file integrity, and acquisition parameters; and
(vi) stated that voxel spacing, calibration, and the complete file inventory could not be established from the available evidence.

\paragraph{Township-level population density.}
The assessment examined whether the response:
(i) identified the temporal, geographic, and township-level scope of the data;
(ii) identified the two GeoTIFF files as primary data;
(iii) recognized the \texttt{TFW}, \texttt{AUX.XML}, and \texttt{OVR} files as sidecars;
(iv) recommended preserving the association between each raster and its sidecars;
(v) proposed checks for georeferencing, CRS, readability, file counts, and consistency; and
(vi) avoided assuming nationwide completeness, spatial resolution, NoData values, or an unsupported distinction between the two rasters.

\paragraph{Dialogue evaluation tables.}
The assessment examined whether the response:
(i) recognized that the dataset contains seven table images rather than a raw dialogue corpus;
(ii) identified \texttt{Table1--5} and \texttt{Table A1--A2};
(iii) distinguished examples and statistics from the two task leaderboards;
(iv) explained that structured values require image inspection, OCR, or manual transcription;
(v) recommended checking all seven files and validating the extracted values; and
(vi) avoided inferring machine-readable records or values not visible in the images.

\paragraph{Weibo rumor-event mapping.}
The assessment examined whether the response:
(i) identified the 4,664 labeled rumor and non-rumor events;
(ii) identified \texttt{events.txt} as containing event identifiers, binary labels, and post identifiers;
(iii) identified \texttt{posts.zip} as containing event-organized JSON posts;
(iv) explained the mapping between event identifiers and the corresponding post files;
(v) recommended checks for label validity, archive extraction, record counts, and referential completeness; and
(vi) avoided assuming exact delimiters or JSON fields without inspecting the files.

\end{document}